\documentclass[conference]{IEEEtran}
\IEEEoverridecommandlockouts
\usepackage{cite}
\usepackage{amsmath,amssymb,amsfonts}
\usepackage{algorithmic}
\usepackage{graphicx}
\usepackage{textcomp}
\usepackage{xcolor}
\def\BibTeX{{\rm B\kern-.05em{\sc i\kern-.025em b}\kern-.08em  T\kern-.1667em\lower.7ex\hbox{E}\kern-.125emX}}

\begin{document}

\title{An Image Forensic Technique Based on JPEG Ghosts}    

\author{\IEEEauthorblockN{Divakar Singh}
\IEEEauthorblockA{
Indian Institute of Technology\\
Jammu, India \\
2019pcs0020@iitjammu.ac.in}
\and
\IEEEauthorblockN{Priyanka Singh}
\IEEEauthorblockA{Dhirubhai Ambani Institute of Information \\
and Communication Technology\\
Gandhinagar, Gujarat \\
priyanka\_singh@daiict.ac.in}
\and
\IEEEauthorblockN{Rajat Subhra Chakraborty}
\IEEEauthorblockA{
Indian Institute of Technology\\
Kharagpur, India \\
rschakraborty@cse.iitkgp.ac.in}

}
\maketitle

\begin{abstract}
The unprecedented growth in the easy availability of photo-editing tools has endangered the power of digital images. An image was supposed to be worth more than a thousand words, but now this can be said only if it can be authenticated or the integrity of the image can be proved to be intact. In this paper, we propose a digital image forensic technique for JPEG images. It can detect any forgery in the image if the forged portion called a ghost image is having a compression quality different from that of the cover image. It is based on resaving the JPEG image at different JPEG qualities, and the detection of the forged portion is maximum when it is saved at the same JPEG quality as the cover image. Also, we can precisely predict the JPEG quality of the cover image by analyzing the similarity using Structural Similarity Index Measure (SSIM) or the energy of the images. The first maxima in SSIM or the first minima in energy correspond to the cover image JPEG quality. We created a  dataset for varying JPEG compression qualities of the ghost and the cover images and validated the scalability of the experimental results.We also, experimented with varied attack scenarios, e.g. high-quality ghost image embedded in low quality of cover image,  low-quality ghost image embedded in high-quality of cover image, and ghost image and cover image both at the same quality.  The proposed method is able to localize the tampered portions accurately even for forgeries as small as $10 \times 10$ sized pixel blocks. Our technique is also robust against other attack scenarios like copy-move forgery, inserting text into image, rescaling (zoom-out/zoom-in) ghost image and then pasting on cover image.
\end{abstract}





\begin{IEEEkeywords}
forgery , JPEG compression , image quality 
\end{IEEEkeywords}

\section{Introduction}
\label{intro}
We are in an era where technology is advancing at a boost-up rate, and with every such boom comes a curse. In this case, the advancement of technology provides anyone the ease to edit/manipulate the image leading to a resultant forged image that is hard to trace, causing a loss of integrity of the image. Nowadays, many fake images are created which can be harmful in many possible ways. In today's digital era, the authenticity of digital data is a prime concern. Image editing software like Adobe Photoshop, Fotor, etc. are readily available which enables the creation of more and more fake images. Some of the areas where image forgery causes irreparable damage are the banking domain (account fraud, cheque frauds), and identity theft. \par

Although many schemes have already been proposed in the literature that addresses various image forgeries still we keep viewing instances of forgeries, e.g. a duplicate move fabrication where a gathering of warriors was copied to cover an image of George Bush \cite{Bush}. A doctored image of a Malaysian politician Jeffrey Wong Su showing him as a knight by the Queen of England in July 2010 \cite{knight}. Another recent news was of a photo shared on Facebook in 2020 to falsely claim that the people in this photo were coronavirus victims in China, but in reality it was a photograph of an art project in Germany in 2014 \cite{art} and many more \cite{news} With high-end editing technologies, it has become really challenging to keep pace with the kind of possible forgeries and their revelation.   \\
Digital image forgery means manipulating a digital image to conceal some meaningful or useful information that would otherwise be conveyed by the image. There are several cases when it is difficult to identify the manipulated/adulterated region of the image. The detection of a forged image is driven by the need for authenticity. Recent advances in digital forensics have given rise to many techniques for detecting photographic tampering. These include techniques for detecting cloning \cite{fridrich2003detection}, \cite{popescu2004exposing}, splicing \cite{ng2004model}, resampling artifacts \cite{avcibas2004classifier,popescu2005exposing}, color filter array aberrations \cite{popescu2005exposing}, sensor noise pattern \cite{lukavs2006detecting}, chromatic aberrations \cite{johnson2006exposing}, and lighting inconsistencies \cite{johnson2007exposing}. Advanced methods like detecting image manipulation
based on edge detection and faster R-CNN \cite{wei2019developing}, image splicing localization using a Multi-task Fully Convolutional Network (MFCN) \cite{salloum2018image} and automatic JPEG ghost detection \cite{azarian2016automatic} were recently proposed.    \par


Methods proposed to detect forgery can be broadly categorized as the active methods and the passive methods. 
In the active approach, certain information is added into an image during the creation of a digital watermark \cite{jarusek2019photomontage}. In the passive method, there is no requirement of active data for authentication of the picture. \par

In this paper, we propose a passive method to detect image forgery based on the difference in the JPEG qualities of the forged portion and rest of the image. The original image is referred to as the cover image and the forged portion is called as a ghost image. The localization of the forged portions is done based on resaving the forged composite image at different image qualities and then finding the range of JPEG qualities where the detection of forgery is maximum. We have experimented with varied combinations of ghost and cover image qualities and also contributed a dataset for scalable testing. The experiments are carried out for varied attack scenarios and analysis of forgery localization done using SSIM and energy of the difference image. Following are the key contributions:

\begin{itemize}
    \item Use of YCbCr color space: For detection of forged portion, we are using a YCbCr color space. In YCbCr color space, the luminance and chrominance components are separated and so it helps the Human Visual System (HVS) to better localize the forged portions. 
    
    \item Dataset: We have constructed a sufficiently large dataset for the forgery detection based on JPEG qualities. A composite image that is
    a combination of ghost and cover image quality is saved at different JPEG qualities in the range of 40 to 100. We tried to cover the maximum combinations of JPEG qualities for  the ghost and the cover image.  
    
    \item Varied attack scenarios: We
    experimented the proposed method with a variety of scenarios to check the robustness of the scheme. To be specific, combinations of high quality ghost-low quality cover, low quality ghost-high quality cover, equal JPEG quality for ghost and cover for copy-move forgeries. Also, we considered forgery like inserting text into images, rescaling (Zoom-out/Zoom-in) of images and ability to detect very small forged ghost portions.
    \item SSIM and energy of image: We analyzed the forgery detection results using SSIM and differy energy graphs and found that the cover image quality can be predicted from these plots. The cover image quality corresponds to first maxima in SSIM plot and first minima in energy plot.  
    
\end{itemize}

The rest of the paper is organized as follows. Section \ref{sec:related} discusses the related work. The proposed method is discussed in section \ref{sec:proposed}. Various experiment scenarios are discussed in \ref{sec:results} along with the details of the dataset. Conclusion and future directions are outlined in \ref{sec:conclusion}.

\section{Related Work}
\label{sec:related}
To detect image forgery, several schemes have been proposed in the literature. Here, we briefly discuss some of these and their limitations.\\
In \cite{popescu2004exposing}, the authors used a technique related to duplicated image regions where they describe an efficient approach that automatically detects copied areas in a digital image. This technique works by applying a principal component analysis to small fixed-size image blocks to yield reduced representation. The accuracy is, in general, excellent, except for small block sizes and low JPEG qualities. The detection rates are nearly perfect, except for small block sizes and low signal-to-noise ratio (SNR). \\
The authors of \cite{avcibas2004classifier} proposed a method based on the assumption that in creation of doctored images, there is always some processing/operations that is done on the images which give rise to measurable distortions in the image properties. They utilized these distortions to classify the images into original verses processed or doctored images. The method was limited to detect forgeries in image regions of dimension at least $100 \times 100$ pixels, not below that which was crucial to many sensitive applications.   \\
 In \cite{lukavs2006detecting}, the authors presented a technique for image forgery detection by checking any distortion in the underlying photo response non-uniformity (PRNU) pattern of the image. PRNU is a unique pattern that can be associated to a specific camera. Any images that are clicked from that camera will have this underlying pattern. Even if two cameras are of make same and model, they will have different PRNU patterns. They proposed their scheme for scenarios when the camera that clicked the photo is available or other images clicked from the same camera are available. They also investigated their scheme for various image processing operations such as lossy compression or filtering and how it influences the ability to verify image integrity. \par
 
The approach in \cite{luo2007novel} proposed a scheme exploiting the symmetry of the blocking artifacts in JPEG images to detect tamper for a cropped and re-compressed image. They derived a blocking artifact characteristic matrix (BACM) for the JPEG images. In case of forgery, the regular symmetry of the BACM gets distorted and this can be used to validate the integrity of the images. Representation features from the BACM was used to train a support vector machine (SVM) classifier and  recognize whether the image is an original JPEG image or it has been cropped from another JPEG image and re-saved. \par

Another image forgery detection was proposed based on statistical correlations that appear in case of forgery \cite{popescu2005exposing}. Whenever a forger wants to make an impercetible forged image, she tries to stretch, resize or rotate the spliced portions of an image to fit properly into another image. This attempt to resample the forged image on a new sampling lattice introduces specific correlations. These correlations can be detected to authenticate or validate the integrity of the image and used for automatic detection of forged portion. This scheme was applicable only to the uncompressed TIFF, JPEG, and GIF images with minimal compression.\par
Another image forgery detection scheme based
on Faster R-CNN model was proposed in \cite{wei2019developing}. 
They combined Laplacian of Gaussian (LoG) operator
and Prewitt operator to detect edges and perform an end-to-end training. It gave satisfactory results for images manipulated  with the addition of Gaussian white noise, Gaussian smoothing,  and JPEG compression. However, it failed for  post-processed images like smoothing the boundary traces between forged and unforged regions. \par

Zhou et al. \cite{zhou2018learning}, proposed a two-stream Faster R-CNN network to extract features that can help distinguish between forged and authentic images. First stream was based on RGB that extracted features based on strong contrast difference, unnatural tampered boundaries, etc. The second was a noise stream focused on leveraging the noise features and chalk out any inconsistency between the  authentic and the tampered regions. The fusion of these streams was then coupled with a bilinear pooling layer to further incorporate spatial co-occurrence of these two modalities. It outperformed many state-of-the-art techniques for NIST Nimble 2016, COVER, CASIA, and the Columbia datasets. \par

Salloum et al. \cite{salloum2018image} proposed another tampering localization technique  by developing a framework that can learn boundaries of the spliced region, and the ground truth mask.  The authors used a Multi-task Fully Convolutional Network for their experiments and found satisfactory results. \par

In \cite{azarian2016automatic}, an algorithm based on SE-MinCut segmentation was proposed to extract the ghost borders. The Bhattacharyya distance was computed to calculate the distance between the original and the tampered regions, which was then fed into the classifier. Although, the automation of ghost detection was solved, but it cannot overcome the problem of low quality ghost-high quality cover image as mentioned in  \cite{farid2009exposing}.
\par
Here, we propose a robust image tamper detection scheme that works efficiently for all possible combinations of JPEG quality for ghost and cover images.

\section{The Proposed Method}
In the proposed method, we present a framework that can be used to verify the integrity of a dubious image as shown in Fig. \ref{prodia}. The method exploits the properties of the JPEG compression to detect the forgery in the images and also localize the tampered regions. Following are the detailed steps:


\begin{figure*}[htbp]
    \centerline{\includegraphics[scale=0.65]{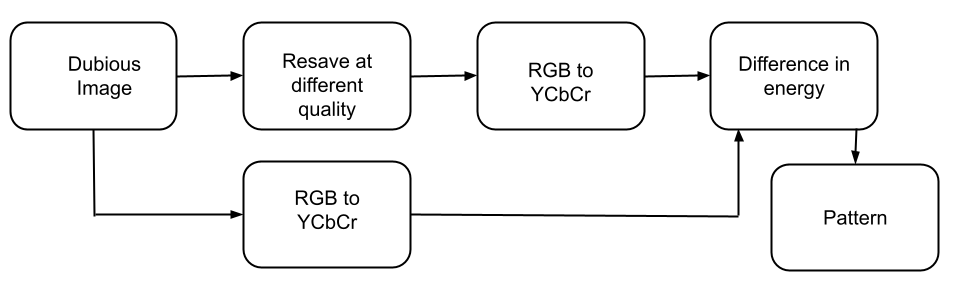}}
    \caption{An overview of the proposed methodology.}
    \label{prodia}
\end{figure*}
\label{sec:proposed}

\textbf{Step 1:} Dubious image: A dubious image is a forged image that is obtained  by copy pasting one portion of the image on another portion within the same image but saved at different JPEG qualities. For example, a central portion $50$ by $50$ of an image with a JPEG quality $60$ called as ghost image, pasted on the cover image originally at quality $90$. Different scenarios of the forgery of the image is discussed in detail in the experiment Section \ref{sec:results}.

\textbf{Step 2:} Resave at different quality: To check the integrity of the dubious image,  it is resaved at different JPEG qualities. We can done this for a range of [30, 100] with  a step size of 2.

\textbf{Step 3:} RGB to YCbCr conversion: RGB color space is changed to YCbCr.Alteration in HVS is more sensitive to brightness changes as compared to color changes. The $Y$ channel holds the luminance information of the image and the color information is contained in $Cb$ and $Cr$ channels. The dubious image $I$ as well the resaved dubious image $I^{q}$ are converted to YCbCr color space using the following: 

\begin{equation}
    \begin{array}{l}
        \mathrm{Y}=0.257 R+0.504 \mathrm{G}+0.098 B+16 \\
        C b=-0.148 \mathrm{R}-0.291 G+0.439 \mathrm{B}+128 \\
        \mathrm{Cr}=0.439 R-0.368 \mathrm{G}-0.071 \mathrm{B}+128
    \end{array}
\end{equation}


\textbf{Step 4:} Difference in energy:  The difference image $D$ is obtained by computing the absolute between the dubious image $I$ and the resaved dubious image $I^{q}$ and amplifying the difference as follows:
\begin{equation}
    D(x, y)= \left [\text{abs}( I(x,y)_{i}- I(x,y)_{i}^{q})\right ]^{3} _{i=1,2,3}
    \label{energy}
\end{equation}

where $I(x, y)_{i}$ and $I(x,y)_{i}^{q}$ represents the pixel value at $(x,y)$ co-ordinates of the $i^{th}$ color channel of the dubious image and resaved dubious image respectively. Here, the $i^{th}$ value represents the $Y$, $Cb$ and $Cr$ channels respectively. 

\textbf{Step 5:} Pattern: To visualize the forged regions clearly, the difference
image $D$ is converted to binary, where the black and white regions represent the untampered and tampered portions of the image.

\section{Experiment Results}
\label{sec:results}
To validate the efficacy of the proposed method, we conducted different experiments on a standard dataset. A variety of attack scenarios were considered for the forgery, keeping the ghost image and the cover image at different JPEG qualities. Based on the results, we can say that the proposed method works efficiently in chalking out the tampered regions even for a high ghost image quality and low cover image quality as well as low ghost image quality and high cover image quality.

\subsection{Dataset}
\label{sec:dataset}

\begin{figure}[htbp]
    \centerline{\includegraphics[scale=0.33]{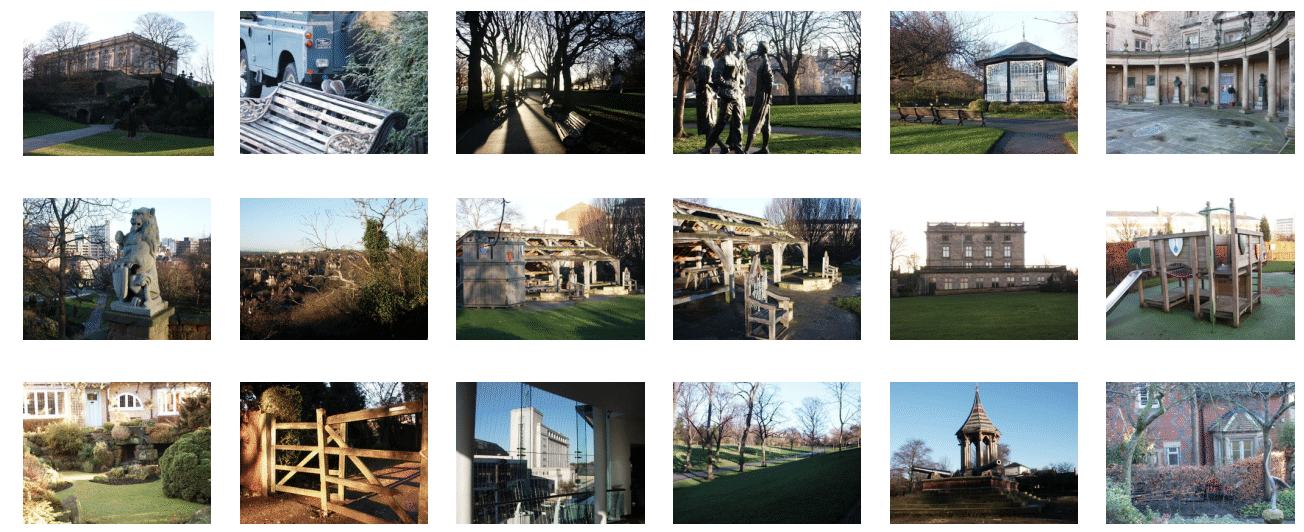}}
    \caption{Some representative images from the UCID dataset.}
    \label{Database}
\end{figure}

The Uncompressed Color Image Database (UCID) is used to validate the experiments for the proposed method \cite{schaefer2003ucid}. It contains high-quality TIFF images, some of which are shown in Fig. \ref{Database}. We have created a database of JPEG images using the images of the UCID database. Firstly, all the 886 images of indoor and outdoor scenes of this database of  size $512 \times 384$ or $384 \times 512$ is converted to JPEG format.  Thereafter, we have made composite images by saving a portion of the image called as ghost image at a different JPEG quality and putting it back into the cover image and resaving the composite image at JPEG quality 100. The range of JPEG qualities that we have considered for the ghost and the cover image is in the range of $\{40, 45, 50, 55, \ldots \ldots, 85, 90, 95, 100 \}$. Combinations of ghost-cover image are organized into multiple folders as follows: 

We have 11 folders $F_{{c_{i}}{g_{j}}}$, where $c$ and $g$ represent the cover image and ghost image with $i$ and $j$ as their JPEG qualities. The values range for $i$ and $j$ falls into  $i = \{40, 45, 50, 55, \ldots, 85, 90, 95 \}$ and $j$ = $\{40, 45, 50, 55, \ldots, 85, 90, 95, 100 \}$. Each folder $F_{c_{i}\_g_{j}}$ again contains 12 sub-folders $Z_{c_{i}\_g_{j}}$, where $c_{i}$ is kept constant and quality of ghost $g_{j}$ is varied for all possible combinations. Hence, each sub-folder contains $886 \times 12$ composite images. Every image in the sub-folder is named as $xx\_yy\_zz.jpg$, where xx represents the image number, yy implies the ghost image quality, and $zz$ communicates the the cover image quality. In total, it contributes a dataset of $886 \times 12 \times 11$ = $1,16,952$ images. 

Apart from this, we also considered attacks like inserting text into images, re-scaling the ghost images. For scale up attack scenarios, we increased $2$ to $3$ times and for scale down, we reduced the ghost image upto $1.5$ times. Also, to test the detection accuracy of the forgery, we tested for ghost image size as small as $10 \times 10$ pixels.


\subsection{Experiment Scenarios}

The experiments were conducted on a machine with Intel(R) Core(TM) i7-10750H CPU @2.60GHz, 64-bit processor, and 16GB RAM with Nvidia GeForce RTX 2060 on MATLAB R2020b. We have considered varied experiment scenarios for maximum possibilities of attack scenarios. For ghost image and cover image composite, we tested all  combinations i.e. high quality of ghost image and low quality of cover image, low quality of ghost image and high quality of cover image, and ghost image and cover image both at the same quality. Other attack scenarios also like copy-move forgery, inserting text into image, rescaling (zoom-out/zoom-in) ghost image and then pasting on cover image.
We also checked the proposed method for it's ability to detect even very small portions of forgery i.e. considering $50 \times 50$, $40 \times 40$ and going upto $10 \times 10$ pixel-sized forgeries.  

The details of each of the experiment scenario along with the supporting results are briefed as follows: 

\begin{enumerate}

    \item The first scenario involves having a ghost image of higher image quality concerning that of the cover image. Results for some images considering it are shown in Fig. \ref{HighLow}.
    This scenario arises when we download the images from social media sites or Apps that compresses the image, then the quality of this image lower than the real camera image. The attacker makes the forged image after downloading the image from social media application or compressed image.
    
    \begin{enumerate}
        \item In Grass field image, Cover image quality-65 and ghost image quality-85 of size $64\times64$ inserted at coordinate  (190, 60).
        \item In stair image, Cover image quality-90 and ghost image quality-55 of size $64\times64$ inserted at coordinate  (190, 60).
    \end{enumerate}


\begin{figure}[htbp]
    \centerline{\includegraphics[scale=0.15]{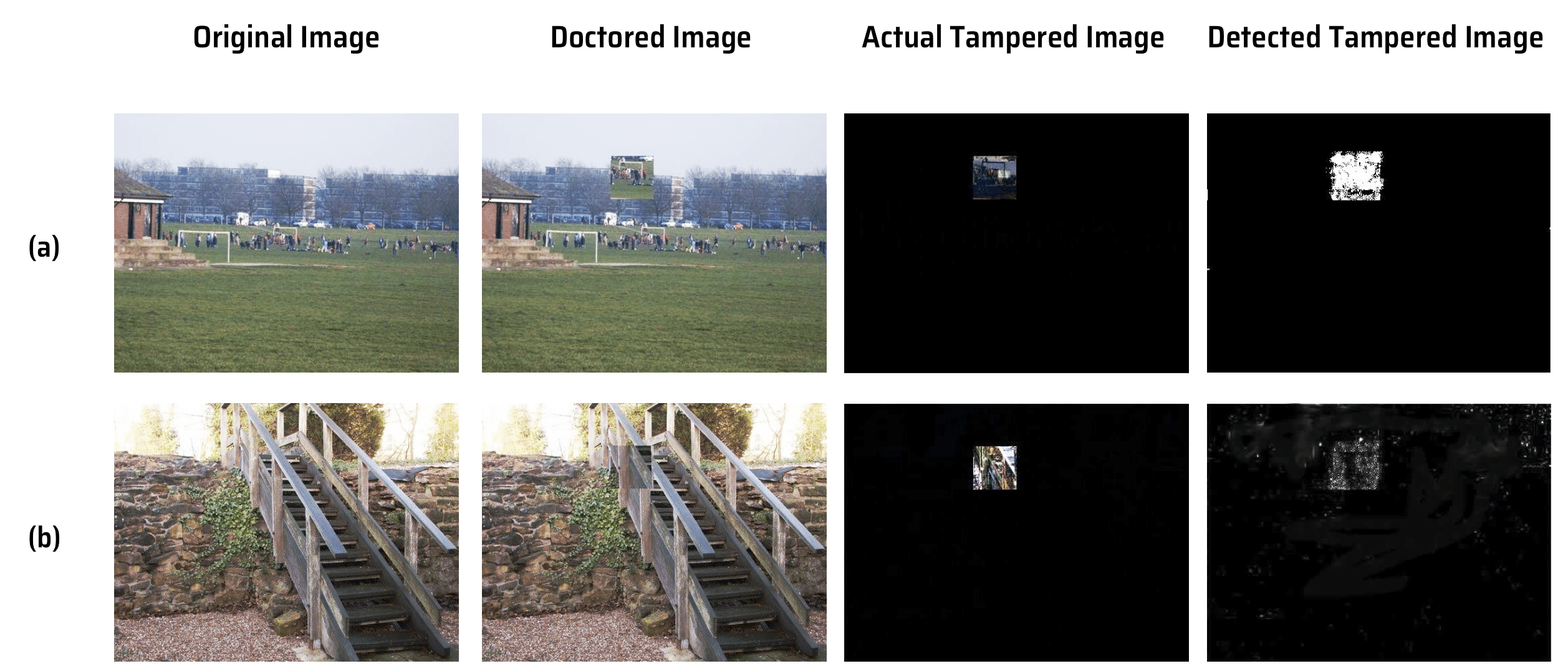}}
    \caption{Results of high quality ghost image inserted into low quality cover image (a) Grass field, (b) Stair.}
    \label{HighLow}
\end{figure}

\item The second scenario involves having the ghost image of a lower image quality when compared to that of the cover image. Results for some images considering it are shown in Fig. \ref{LowHigh}. 
    
    \begin{enumerate}
        \item In man image, Cover image quality-90 and ghost image quality-70 of size approx. $89\times245$ inserted at coordinate  (190, 60).
        \item In house image, Cover image quality-70 and ghost image quality-50 of size $64\times64$ inserted at coordinate  (190, 60).
    \end{enumerate}


\begin{figure}[htbp]
    \centerline{\includegraphics[scale=0.25]{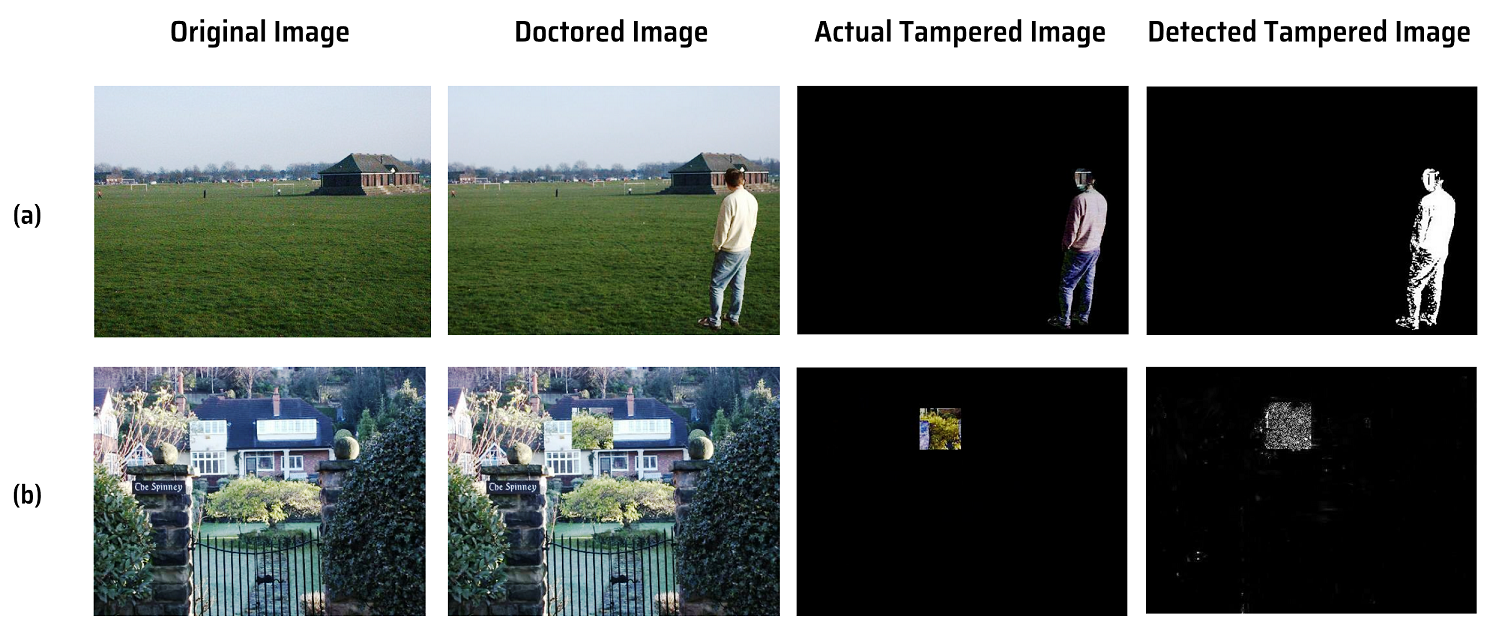}}
    \caption{Results of low quality ghost image inserted into high quality cover image (a) Man, (b) House.}
    \label{LowHigh}
\end{figure}

\item In the third scenario, both the cover image and the ghost image have the same quality. It may be both ghost and cover image compressed at some quality. Results for some images considering it are shown in Fig. \ref{Samequality}. 

    \begin{enumerate}
        \item In wall image, Cover image quality-75 and ghost image quality-75 inserted at coordinate  (190, 60) of size $64\times64$.
        \item In tree image, Cover image quality-45 and ghost image quality-45 inserted at coordinate  (190, 60) of size $64\times64$.
    \end{enumerate}


\begin{figure}[htbp]
    \centerline{\includegraphics[scale=0.15]{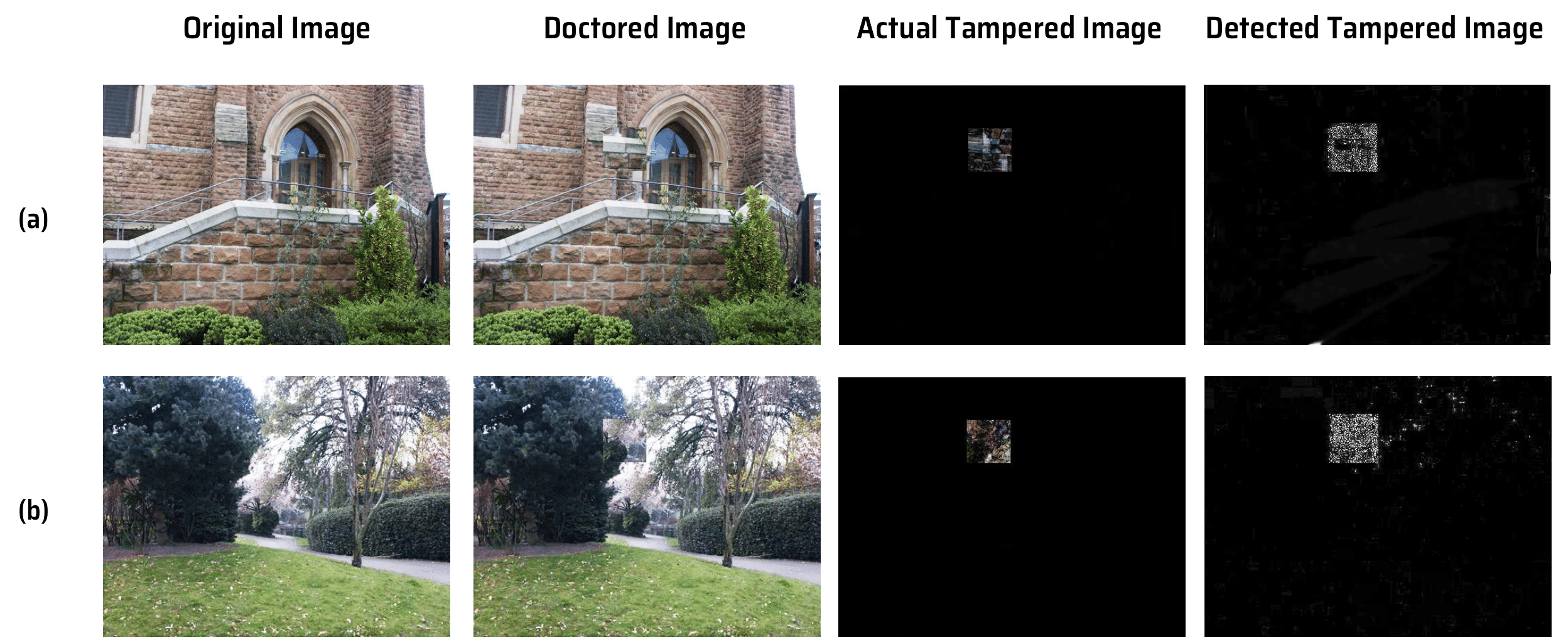}}
    \caption{Results of both the image (Ghost and cover image) quality is same. : (a) Wall , (b) Tree.}
    \label{Samequality}
\end{figure}

\item In this scenario, the ghost image and cover image are the same quality because it is originally a part of the cover image and copied and pasted in the same cover image.  Results for some images considering it are shown in Fig. \ref{Copy-Move}.  

    \begin{enumerate}
        \item In wall image, Cover image quality-45 and ghost image quality-45 of size $64\times64$ inserted at coordinate  (190, 60).
        \item In field image, Cover image quality-55 and ghost image quality-55 of size $30\times68$ inserted at coordinate  (398, 170).
    \end{enumerate}


\begin{figure}[htbp]
    \centerline{\includegraphics[scale=0.15]{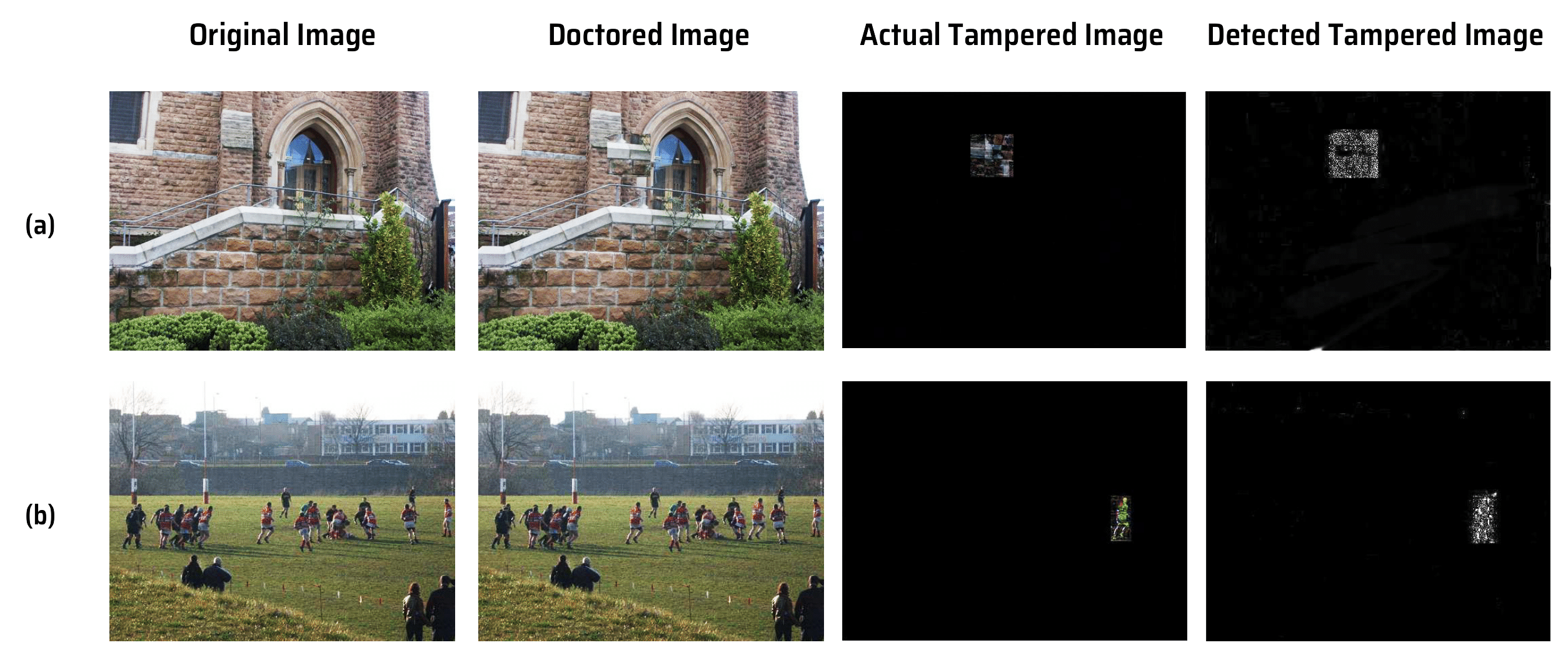}}
    \caption{Results of copy and paste the portion of image into same image. : (a) Wall, (b) Field.}
    \label{Copy-Move}
\end{figure}

\item In this scenario, Inserting text into the cover image shows that the organisation's property or person. In this scenario, we cover the inserting text into image forgery detection, as shown in Fig. \ref{InsertingText}. 

    \begin{enumerate}
        \item In the road image, the cover image's quality is 85, and it contains text, i.e., "Hello MATLAB!" at coordinate (197,243) of size 10.
        \item In the glass image, the cover image's quality is 75, and it contains text, i.e., "Abcd" at coordinate (248,192) of size 14.
    \end{enumerate}


\begin{figure}[htbp]
    \centerline{\includegraphics[scale=0.17]{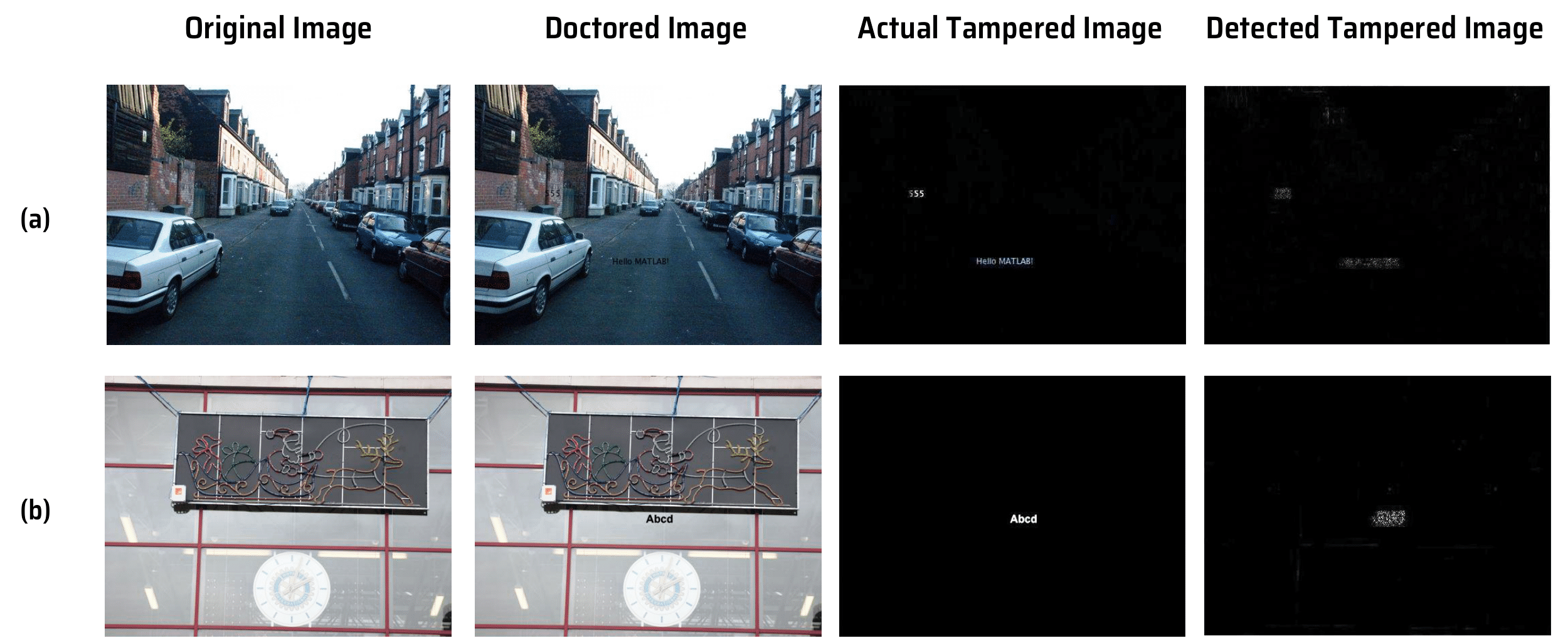}}
    \caption{Results of inserting text into image. : (a) Road, (b) Glass.}
    \label{InsertingText}
\end{figure}

\item The sixth scenario involves taking portions of the cover image as the ghost images and resizing them (zoom out/in). Zoom out/in means enlarging/reducing the size of a picture in a sense and pasting them onto the cover image. Results for some images considering it are shown in Fig. \ref{RescaledImage}.

    \begin{enumerate}
        \item In the house image, the cover image's quality is 65, and it contains a windows picture in the zoom-out form at coordinate (359,102) of size $62\times37$ to $105\times76$.
        \item In the tiger image, the cover image's quality is 40, and it has a copy of the tiger in the zoom-in form at coordinate (352,183) of size $240\times160$ to $148\times87$.
    \end{enumerate}


\begin{figure}[htbp]
    \centerline{\includegraphics[scale=0.15]{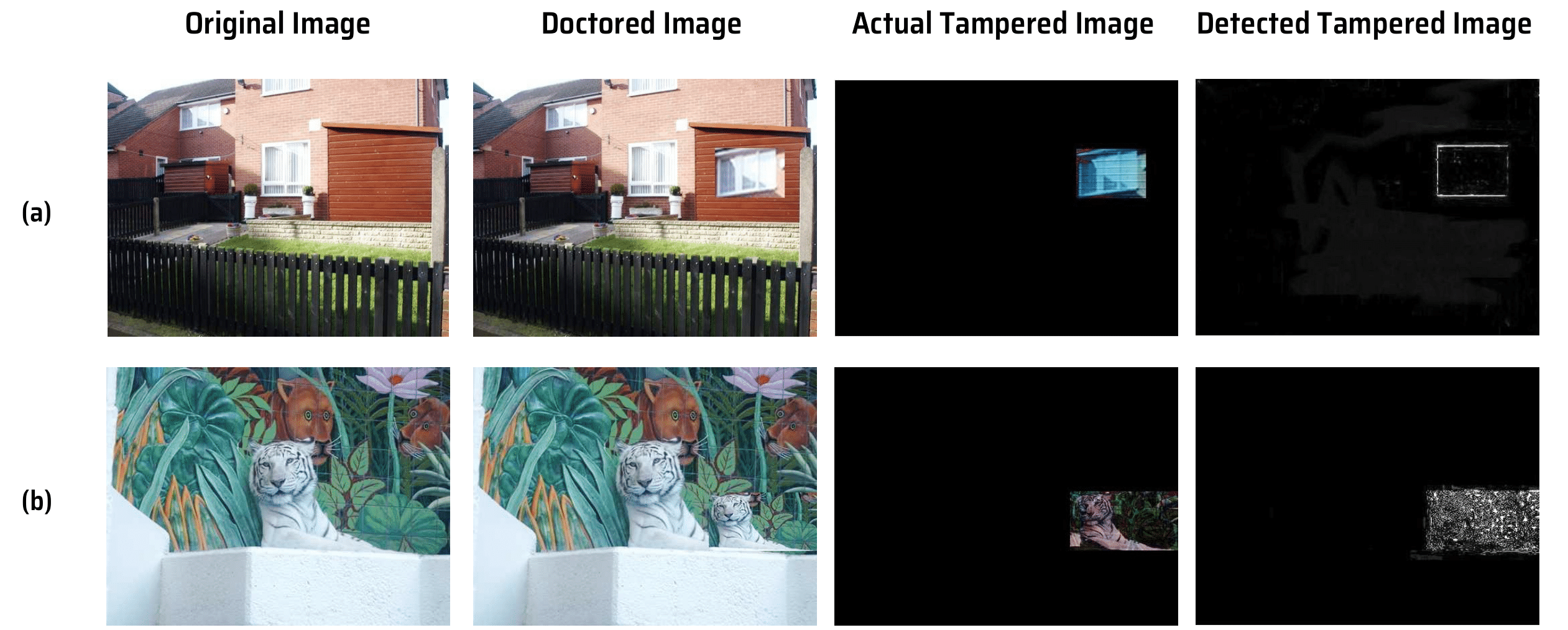}}
    \caption{Results of Rescale (Zoom-in / Zoom-out) an object in the image. : (a) House , (b) Tiger.}
    \label{RescaledImage}
\end{figure}

\item In this scenario.  the ghost image size is kept 10x10, 20X20, 30X30, 40x40, 50x50, 60x60 pixels in the cover image to detect the adulteration. Results for some images considering it are shown in Fig. \ref{size of Ghost image 10,and 30}.  In this experiment, we show how small the size of the ghost image can be detected / visible.
    
    \begin{enumerate}
        \item In the sliding image, the cover image's quality is 85, and ghost image quality 65 at coordinate (29,9) of size $10\times10$.
        \item In the wood-house image, the cover image's quality is 85, and ghost image quality 65 at coordinate (89,28) of size $30\times30$.
    \end{enumerate}


\begin{figure}[htbp]
    \centerline{\includegraphics[scale=0.17]{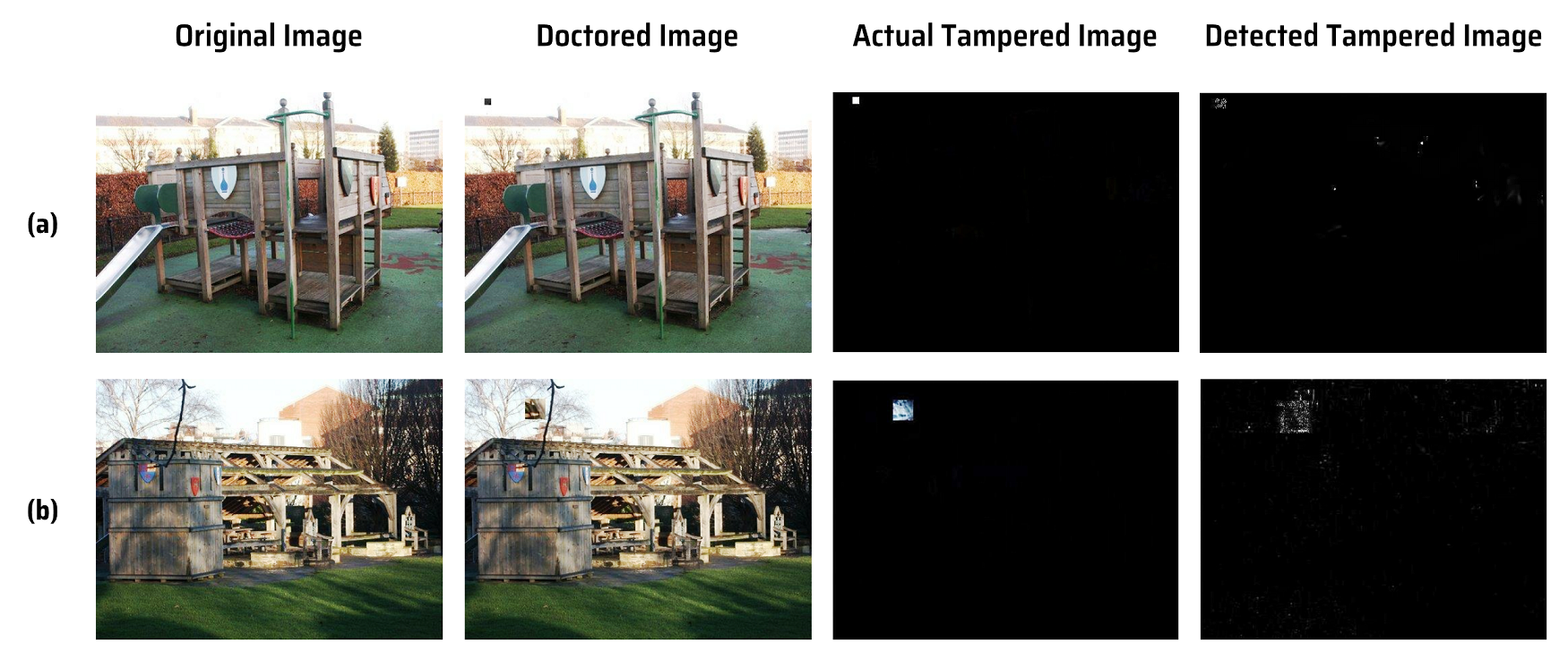}}
    \caption{Results of JPEG ghost block size $10\times10$ and $30\times30$. : (a) Sliding, (b) wood-house.}
    \label{size of Ghost image 10,and 30}
\end{figure}

\end{enumerate}

We analyzed the different combinations of ghost and cover image by varying the JPEG quality for the resaved composite images. Based on the experiments and testing with a sufficiently large dataset, we found that the detection of the forged regions is good for a range of JPEG quality compared to the entire range and possible combinations for ghost and cover images. We exploited SSIM \cite{hore2010image} to measure the similarity between a forged image $I$ and it's corresponding resaved version $I^{q}$ at a quality $q$ using following: \par

The SSIM formula is based on three comparison measurements between the samples of $I$ and $I^{q}$: luminance (l), contrast (c) and structure (s). The individual comparison functions are:

\begin{equation}
 \operatorname{SSIM}(I, I^{q})=[l(I, I^{q})] \cdot[c(I, I^{q})] \cdot[s(I, I^{q})] 
\end{equation}
where
\begin{equation}
    l(I, I^{q}) =\frac{2 \mu_{I} \mu_{I^{q}}+C_{1}}{\mu_{I}^{2}+\mu_{I^{q}}^{2}+C_{1}}
\end{equation}

\begin{equation}
     c(I, I^{q}) =\frac{2 \sigma_{I} \sigma_{I^{q}}+C_{2}}{\sigma_{I}^{2}+\sigma_{I^{q}}^{2}+C_{2}}
\end{equation}

\begin{equation}
    s(I, I^{q}) =\frac{\sigma_{I I^{q}}+C_{3}}{\sigma_{I} \sigma_{I^{q}}+C_{3}}
\end{equation}


Here, $\mu_{I}$ is the average of $I$ , $\mu_{I^{q}}$ is the average of $I^{q}$ , $\sigma_{I}^{2}$ is the variance of $I$ , $\sigma_{I^{q}}^{2}$ is the variance of $I^{q}$, $\sigma_{I I^{q}}$ is the covariance of $I$ and $y$, $c_{1}=\left(k_{1} L\right)^{2}$, and  $c_{2}=\left(k_{2} L\right)^{2}$ are two variables to stabilize the division with weak denominator, $L$ is the dynamic range of the pixel-values, $k_{1}=0.01$, $k_{2}=0.03$, and $c_{3}= \frac{c_{2}}{2}$.

\begin{figure}[htbp]
   \centerline{\includegraphics[scale=0.23]{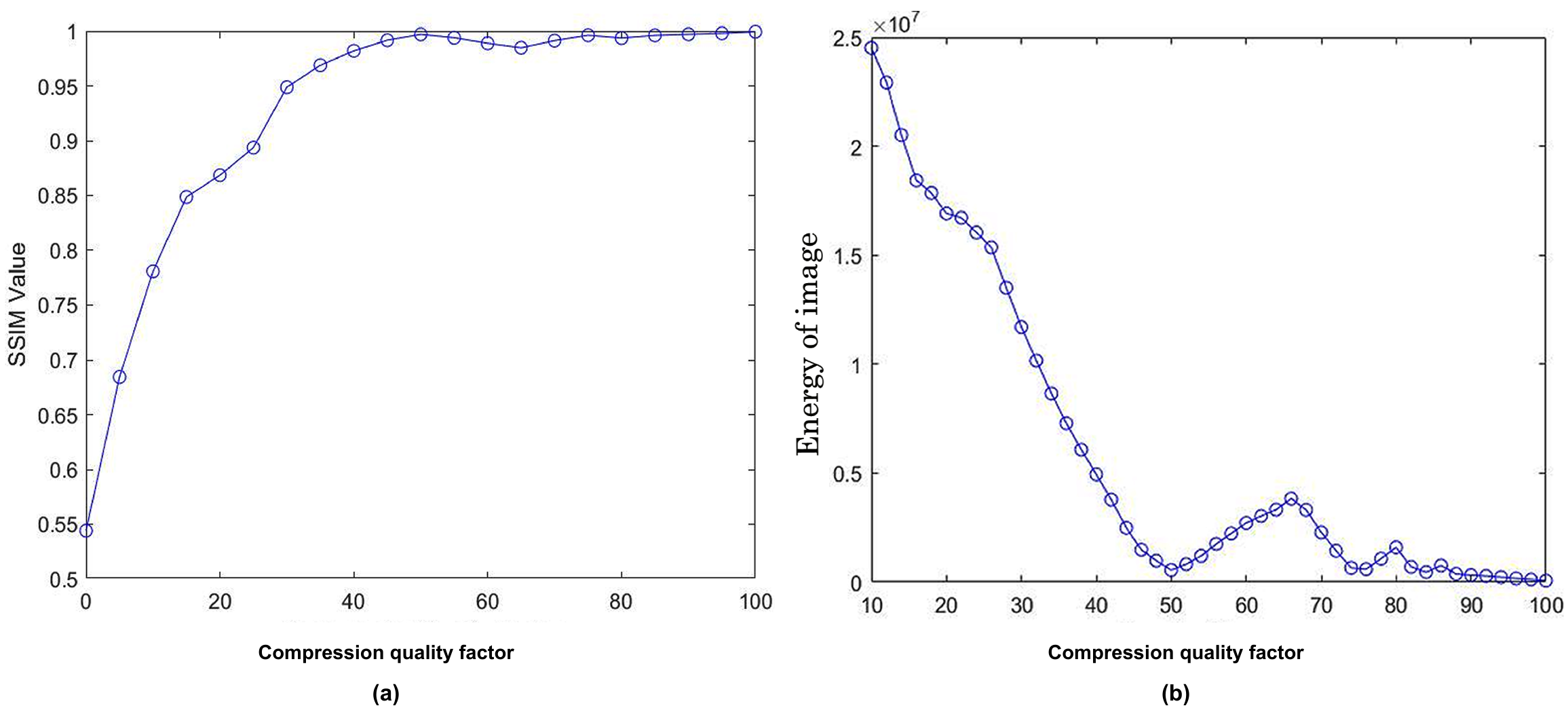}}
    \caption{(a) SSIM Graph (b)Energy Graph}
    \label{ssim_energy}
\end{figure}

By analyzing plots of "SSIM value"  versus "compression quality factor" of the resaved image, we found that the first maxima occurs at values near the original quality of the cover image as shown in Fig. \ref{ssim_energy} (a). Here, the first maxima occurs at 50 JPEG quality which was also the quality of the the cover image.

We performed another analysis for the experiment scenarios based on the difference in energy of the forged and it's resaved version using Equation \ref{energycom}. 

To calculate energy of image-
\begin{equation}
    E(x,y)=\sum_{d=1}^{dim} \sum_{x=1}^{rows} \sum_{y=1}^{col}  D(x,y,d)
\label{energycom}
\end{equation}
where
    $E(x,y)$ represents the sum of the amplified pixel values $D(x,y)_i$ of the difference image obtained in 
    Equation \ref{energy}.

The first minima in plots of "energy of image"  versus "compression quality factor" correspond to the quality of the cover image quality. For instance, the first minima in  Fig. \ref{ssim_energy} (b) occur at compression quality 50, which again corresponds to the quality of the cover image. Hence, this cross-verifies the result that we obtained using SSIM plots.

\subsection{Comparison Table}
A comparative study of the proposed scheme with the other state-of-the-art tamper detection approaches based on various criteria has been tabulated in Table \ref{criteria} and \ref{attackscenarios}. A few of the criteria are described as follows:

\begin{table}[htbp]
\caption{Comparison based on various criteria}
\label{criteria}
\centering
\tiny
\begin{tabular}{|l|c|c|c|c|}
\hline
 &
 \textbf{\begin{tabular}[c]{@{}c@{}}Blind\\  authentication\end{tabular}} &
  \textbf{\begin{tabular}[c]{@{}c@{}}Is detection \\ possible with \\ just one image?\end{tabular}} &
  \textbf{\begin{tabular}[c]{@{}c@{}}Can detect \\  very minute\\  forgeries \end{tabular}} &
  \textbf{\begin{tabular}[c]{@{}c@{}}Can predict \\ quality of \\ original image\end{tabular}} \\ \hline
Wu et al. [2020] \cite{wu2020sharing}      & No  & Yes & Yes & No  \\ \hline
Thai et al. [2017] \cite{thai2016JPEG}         & Yes & Yes & Yes & Yes \\ \hline
Hou et al. [2014] \cite{hou2014hue}           & Yes & No  & No  & No  \\ \hline
Chierchia et al. [2014] \cite{chierchia2014bayesian}  & Yes & Yes & No  & No  \\ \hline
Agarwal et al. [2017] \cite{agarwal2017photo}       & Yes & Yes & -   & No  \\ \hline
Azarian et al. [2016] \cite{azarian2016automatic}  & Yes & Yes & Yes & Yes \\ \hline
Kumawat et al. [2020]  \cite{kumawat2020robust}   & Yes & Yes & -   & No  \\ \hline
Pun et al. [2015] \cite{pun2015image}          & Yes & Yes & Yes & No  \\ \hline
Chen et al. [2019] \cite{chen2019rotational} & Yes & Yes & Yes & No  \\ \hline
Mullan et al. [2019] \cite{mullan2019forensic} & No & Yes & - & No  \\ \hline
Bhardwaj et al. [2018] \cite{bhardwaj2018JPEG} & Yes & Yes & - & Yes  \\ \hline
Proposed Scheme    & Yes & Yes & Yes & Yes \\ \hline
\end{tabular}%
\end{table}

\begin{table}[htbp]
\caption{Comparison based on various attack scenarios}
\label{attackscenarios}
\centering
\tiny
\begin{tabular}{|l|c|c|c|c|}
\hline
 & \textbf{JPEG Quality} & \textbf{Copy-move} & \textbf{Text Insertion} & \textbf{Rescale} \\ \hline
Wu et al. [2020] \cite{wu2020sharing}      & No  & No  & No  & No  \\ \hline
Thai et al. [2017] \cite{thai2016JPEG}           & Yes & Yes & No  & No  \\ \hline
Hou et al. [2014] \cite{hou2014hue}               & Yes & No  & No  & No  \\ \hline
Chierchia et al. [2014] \cite{chierchia2014bayesian}   & No  & Yes & No  & No  \\ \hline
Agarwal et al. [2017] \cite{agarwal2017photo}        & Yes & Yes & Yes & No  \\ \hline
Azarian et al. [2016] \cite{azarian2016automatic}    & Yes & No  & No  & No  \\ \hline
Kumawat et al. [2020]  \cite{kumawat2020robust}      & No  & No  & No  & No  \\ \hline
Pun et al. [2015] \cite{pun2015image}          &  -  & Yes & No  & No  \\ \hline
Chen et al. [2019] \cite{chen2019rotational} & No  & Yes & No  & No  \\ \hline
Mullan et al. [2019] \cite{mullan2019forensic} & No & Yes & -  & No  \\ \hline
Bhardwaj et al. [2018] \cite{bhardwaj2018JPEG} & - & - & No & No  \\ \hline
Proposed Scheme    &Yes  &Yes  & Yes & Yes \\ \hline
\end{tabular}%
\end{table}

\begin{itemize}
  \item \textbf{Blind authentication}: If the integrity of an image can be checked just based on the content of the given image without any additional information or attributes, it is termed as blind authentication.

  \item \textbf{Is detection possible with just one image?}: This criteria is based on whether the forgery detection can be done with just a single available image or a bunch of images are needed to detect the forged regions. For example, detection based on Photo Response Non-Uniformity (PRNU) of a camera. 
  
  \item \textbf{Can the technique detect very minute forgeries?}: This criteria is based on what level of forgery detection can be done by a scheme. We are referring to forgeries less than $10\times10$ sized pixels as the minute forgeries. 
  
  \item \textbf{Can the technique predict quality of the original image?}: This criteria is based on whether an algorithm can predict the JPEG  quality of the original cover image or not. 
  
  The proposed scheme can narrow down this range of possible JPEG quality and can detect in very less iterations.  
  
  \item \textbf{JPEG quality}: In this type of forgery,  the content of an image at JPEG quality $q_{1}$ is copied and pasted into another image of JPEG quality $q_{2}$. This composite image is then resaved at  different JPEG qualities and forgery detection is performed based on these quality differences. 
  
  \item \textbf{Copy-move}: In this type of forgery,  a portion of an image is copied and pasted into another place in the same image to hide other information or falsify the original content of the image. 
  
  \item \textbf{Text insertion}: In this type of attack, some text messages are inserted on top of the original content of the image. 
  
  \item \textbf{Rescale}: In this type of attack, few portions of the original image are rescaled (ZoomIn/ZoomOut) and pasted over to conceal some important information in the original image to falsify the information.
\end{itemize}

From Table \ref{criteria} and \ref{attackscenarios}, we can observe that only two schemes proposed in \cite{wu2020sharing, mullan2019forensic} cannot do the blind authentication. They require additional information. Sharing phase  of \cite{wu2020sharing}, a secret image is encoded into shared bits by polynomial based secret image sharing. Source linkage based on header information of media items allow for easy automation in \cite{mullan2019forensic}. In general, detection of forged regions based on just a single image becomes quite challenging.  Hence it is an important criteria to judge usuability of a proposed scheme. This is possible for all the schemes considered for comparison here except for \cite{hou2014hue}. \cite{hou2014hue} needs to achieve this goal, we use sensor pattern noise from each color channel of untampered images as the ground truth. The level of forgery detection is very crucial for sensitive applications. Forgeries as minute as $10 \times 10$ sized blocks can cause a lot of damage for critical applications. The proposed scheme is capable to detect such minute forgeries. 
Many forgery detection schemes become  computation expensive as they have to repeat  the entire process again and again to detect the forged regions. One such forgery detection is  based on JPEG qualities. If it is possible to predict the original quality of the JPEG image, then this process can be accelerated.  In the proposed scheme, the quality of the original image can be predicted using the SSIM and energy graphs. \cite{thai2016JPEG,hou2014hue,agarwal2017photo,azarian2016automatic} schemes can detect forgery based on difference of JPEG quality in the cover and the ghost image. \par

Varied attack scenarios such as attacks based on JPEG quality, copy-move forgery, insertion of text into images and rescaling a portion for falsifying the original information are considered to evaluate the performance of the proposed scheme. Table \ref{attackscenarios} enlists the attack scenarios. The state-of-the-art schemes are compared based on their robustness towards these scenarios. The proposed scheme can tackle all the aforementioned attacks.  


\section{Conclusion and Future Research Directions}
\label{sec:conclusion}
This paper describes an effective technique for detecting tampering in  JPEG images. Based on the experiments, we have observed that if two different images of different JPEG qualities are combined to obtain a composite image, the possibility of detecting the forgery is quite high. Based on the study in the experiments, the quality of the cover image can be predicted based on the "SSIM" and "energy of image" verses "compression quality factor" plots. \par
The proposed approach helps in reducing a lot of efforts needed to detect the tampered portion as well, as one can directly check the narrow range near the maxima/minima points in the SSIM/Energy plots.  However, if the two combined images have the same JPEG quality, the detection possibility becomes quite low. This is irrespective of the fact whether the images were captured from the same camera device or not. The future directions could be localizing the forgery in images that have undergone multiple compressions. Another research direction could be localizing the forged area of images with multiple forgeries.




\bibliographystyle{IEEEtran} 
\bibliography{cas-refs}

\end{document}